\documentclass[10pt,twocolumn,letterpaper]{article}

\usepackage[pagenumbers]{cvpr} % To force page numbers, e.g. for an arXiv version

\usepackage{multirow}
\definecolor{cvprblue}{rgb}{0.21,0.49,0.74}
\usepackage[pagebackref,breaklinks,colorlinks,allcolors=cvprblue]{hyperref}

\def\paperID{*****} % *** Enter the Paper ID here
\def\confName{CVPR}
\def\confYear{2026}

\title{ScribbleEdit: Synthetic Data for Image Editing with Scribbles and Text}

\author{Anya Ji\thanks{Equal Contribution.}
\qquad
George Ma\footnotemark[1]
\qquad
T\'ea Wright\footnotemark[1]
\qquad
Yiming Zhang\footnotemark[1]
\\
David M. Chan
\qquad
Alane Suhr
\qquad
Somayeh Sojoudi\\
University of California, Berkeley\\
{\tt\small \{anyaji,george\_ma,teaywright,ym\_zhang,davidchan,suhr,sojoudi\}@berkeley.edu}
}

\begin{document}
\maketitle

\begin{abstract}
Recent progress in generative models has significantly advanced image editing capabilities, yet precise and intuitive user control remains difficult. Specifically, users often struggle to communicate both exact spatial layouts and specific semantic details simultaneously. While natural language instructions effectively convey high-level semantics like texture and color, they lack spatial specificity. Conversely, freehand scribbles provide rough spatial boundaries but cannot express detailed visual attributes.
Consequently, achieving precise control requires combining both modalities. However, existing models struggle to jointly interpret abstract scribbles alongside text due to a lack of specialized training data.

In this work, we introduce \textbf{ScribbleEdit}, a large-scale synthetic dataset designed to bridge this gap by combining natural language instructions with freehand scribble inputs for more accurate, controllable edits.
We construct this dataset through a synthetic pipeline that automatically generates source-target image pairs via inpainting, which are then paired with human-drawn scribbles and VLM-generated text instructions.
Using ScribbleEdit, we evaluate and finetune both diffusion-based and autoregressive unified multimodal image editing models. Our experiments reveal that while off-the-shelf models struggle with abstract scribble inputs, finetuning on our synthetic dataset significantly improves their ability to generate spatially aligned and semantically consistent edits.
%We release our dataset and results at: \TODO{dashboard}
\end{abstract}

\section{Introduction and Background}
\label{sec:intro}
Despite progress in image editing algorithms, we are still unable to achieve precise control over edits, primarily due to the limitations of natural language.
% \citep{brooks2023instructpix2pixlearningfollowimage,xu2024inversion,santos2025hands,shi2021learning}
While language instructions are expressive for describing high-level semantics, they often lack spatial specificity, making it difficult to control where and how edits appear. Freehand scribbles provide a complementary solution, offering an intuitive control for users to quickly outline an edit's location and structure. However, because scribbles are inherently abstract and imprecise representations, they are significantly harder for models to interpret and translate into accurate edits.

\begin{figure}[t]
    \begin{minipage}{\columnwidth}
        \hfill
        \includegraphics[width=\columnwidth]{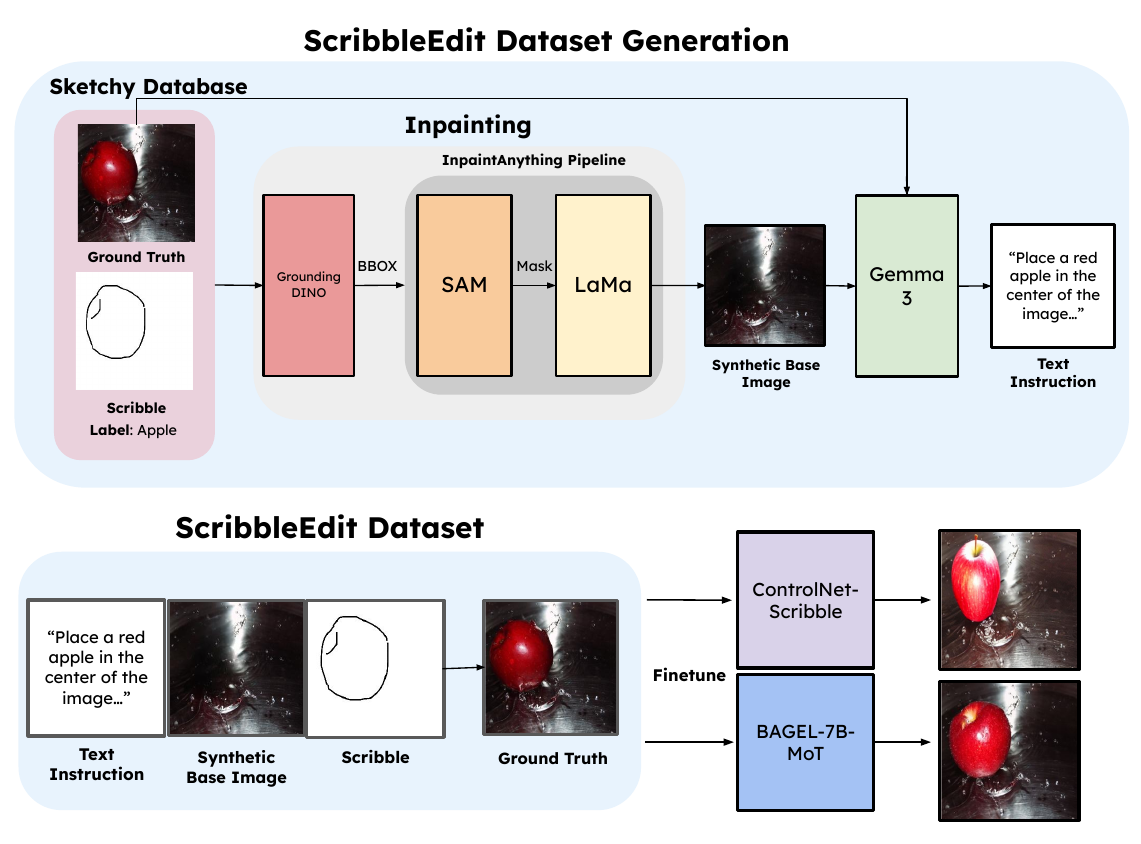}
        \caption{Overview of the ScribbleEdit pipeline for dataset construction and model finetuning.}
        \label{fig:firstpage}
    \end{minipage}
\end{figure}

Mask-based inpainting is a common technique for generating training data for image editing by removing objects and generating textual instructions for reconstruction \citep{wasserman2025paint,bala2024galaxyedit}. However, these approaches typically focus on text-only editing and provide limited spatial guidance. Scribbles offer a lightweight visual signal that captures rough spatial structure while leaving appearance details to the generative model. Unlike precise masks or detailed sketches, scribbles are abstract and imprecise, making them challenging for existing models to interpret reliably. At the same time, they are intuitive and efficient for users to produce, requiring only minimal effort to indicate approximate shape and placement. This makes scribbles a practical interface for controllable image editing while posing a challenge for generative models.

Prior work has explored sketch-guided editing \citep{zeng2022sketchedit,sharma2024sketch,mao2023sketchffusion}, as well as multimodal conditioning for image generation using text and visual inputs such as sketches, edges, or masks \citep{zhang2023adding,mou2024t2i,hu2024instruct}. For image editing, SmartBrush \cite{xie2023smartbrush} conditions object inpainting on both text and shape for semantic and spatial guidance. ReEdit \citep{srivastava2025reedit} further reduces ambiguity by incorporating exemplar images alongside text. Interactive systems such as MagicQuill \citep{liu2025magicquill} further demonstrate the potential of combining multiple visual cues for controllable editing. However, datasets designed specifically for studying scribble- and text-guided image editing remain limited.

To address this gap, we introduce \textbf{ScribbleEdit}, a large-scale synthetic dataset for multimodal image editing that combines scribbles and natural language instructions. Our pipeline constructs paired examples by removing objects from images in the Sketchy Database \citep{sketchy-database} and generating instructions for restoring them, as illustrated in Figure~\ref{fig:firstpage}. 

We evaluate the dataset by training and benchmarking diffusion-based and unified multimodal models under different conditioning settings, including text-only, scribble-only, and joint scribble–text guidance. Our results show that scribble-guided editing remains challenging for current models and finetuning on ScribbleEdit effectively improves models' ability to generate spatially accurate and semantically consistent images based on scribble and text. Overall, our contributions are: (1) ScribbleEdit, a dataset for scribble- and text-guided image editing; (2) an automated pipeline for constructing multimodal editing examples; and (3) a benchmark evaluation of modern image editing models under different conditioning settings.

\section{Dataset Generation}
\label{sec:dataset}

We construct a dataset of 11,775 samples for the object addition task, where the goal is to add a specific object to an image guided by a scribble and a natural language instruction.
Each sample consists of four components: an \textbf{input image} with the object removed, the original photo serving as the \textbf{target image}, a hand-drawn \textbf{scribble} of the object, and a natural language \textbf{instruction} describing the edit.

\paragraph{Source data.}
We subsample 12,500 image-scribble pairs from the Sketchy Database \citep{sketchy-database}, a large-scale collection of photos paired with hand-drawn scribbles contributed by crowd workers.
Each photo depicts a single object from one of 125 categories; every photo is accompanied by between 5 and 18 scribbles (6 on average), each drawn by a different annotator.
We retain samples in which the object occupies a reasonable portion of the image and the scribbles visually correspond to the object's position in the photo.

\paragraph{Object segmentation.}
To obtain object masks, we apply a two-stage pipeline.
First, grounding DINO \citep{grounding-dino} detects the object using the class name as the text prompt to produce a bounding box. 
Second, the box is passed to the Segment Anything Model (SAM) \citep{segment-anything} (ViT-H variant), which produces three mask candidates.
We select the candidate with the highest predicted IoU score.
The selected mask is dialted by 15 pixels using a $15 \times 15$ structuring element to ensure complete coverage of the object boundary.

\paragraph{Input image generation.}
We remove the segmented object from each photo using the LaMa inpainting model \citep{lama}, which fills the masked region with a plausible background.
Because the mask already identifies the object precisely, no text prompt is needed; LaMa reconstructs the background purely from the surrounding image context.
The resulting inpainted image serves as the input to the editing model.
Of the 12,500 sampled photos, 11,775 yield a valid input image after filtering out cases where object detection or segmentation fails.

\paragraph{Instruction generation.}
We generate a natural language instruction for each sample using Gemma~3 (4B) \citep{gemma-3}, an instruction-tuned vision-language model.
The model is given the input image and the target image together with a prompt that specifies the editing operation (e.g., ``Add a \textit{class} to the image.'') and asks the model to produce a single, specific editing command that describes the object's appearance and location.
% Example generated instructions include \textit{``Place a red and yellow striped apple with a stem in the center of the image.''} and \textit{``Add an airplane to the lower-right quadrant, oriented horizontally with a red and white fuselage.''}.

\paragraph{Dataset splits.}
We split the 11,775 samples into training, validation, and test sets containing 9,438, 1,164, and 1,173 unique photos, respectively.
Since each photo is paired with multiple scribbles, to reduce training cost, we sample a subset in which each photo is paired with at most two scribbles, giving 18,876 training, 2,328 validation, and 2,346 test instances. An example is shown in Figure~\ref{fig:firstpage}.

% putting it here so that it stays on page 3
\begin{table*}[t]
\centering
\resizebox{\textwidth}{!}{
\setlength{\tabcolsep}{4pt}
\begin{tabular}{l c ccccccc}
\toprule
\textbf{Model} & \textbf{Inputs} &
\multicolumn{2}{c}{\textbf{Histogram Intersection} $\uparrow$} &
\multicolumn{2}{c}{\textbf{RMSE} $\downarrow$} &
\textbf{CLIP} $\uparrow$ &
\textbf{FID} $\downarrow$ \\
 &  & Entire & BBOX & Entire & BBOX & Entire & Entire \\
\midrule

SD Inpaint + ControlNet-Scribble \cite{zhang2023adding} 
& S+T+M & \underline{0.82} & 0.49 & \textbf{36.10} & 78.76 & 0.81 & 53.63 \\

BAGEL-7B-MoT \cite{deng2025bagel} 
& S+T & 0.63 & 0.53 & 58.13 & 71.55 & 0.68 & 112.92 \\
\midrule
\multirow{3}{*}{Image-Cond.\ ControlNet-Scribble (finetuned)}
& T & 0.68 & 0.49 & 56.94 & 82.12 & \underline{0.87} & \underline{28.90} \\
& S & 0.68 & 0.51 & 55.16 & 78.02 & 0.71 & 89.60 \\
& S+T & 0.68 & 0.49 & 55.65 & 82.10 & \underline{0.87} & \textbf{28.47} \\

\midrule
\multirow{3}{*}{BAGEL-7B-MoT (finetuned)}
& T & \underline{0.82} & \underline{0.70} & 39.76 & 59.83 & 0.81 & 56.96 \\
& S & 0.81 & \underline{0.70} & 36.86 & \underline{56.80} & 0.86 & 40.50 \\
& S+T & \textbf{0.83} & \textbf{0.72} & \underline{36.14} & \textbf{55.77} & \textbf{0.88} & 33.36 \\

\bottomrule
\end{tabular}}
\caption{Quantitative results on our dataset. Inputs denote scribble (S), text instruction (T), and mask (M). \textbf{Bold} indicates the best result in each column, and \underline{underline} indicates the second best.}

\label{tab:editing-results}
\end{table*}

\section{Experiments}

We evaluate baseline and finetuned diffusion-based models and multimodal large language models on multimodal image editing conditioned on three modalities: an original image, a scribble indicating the desired structure, and a natural-language edit instruction. 
The scribble indicates the approximate shape and location of the object to be added. It provides spatial guidance for where and how the new object should appear, while the text instruction specifies semantic attributes such as the object category, color, or style. The model must combine these two complementary signals: the scribble constrains the rough geometry and placement of the edit, while the language instruction defines its visual properties.
Table~\ref{tab:editing-results} reports quantitative results.

\subsection{Models}
We include the following baselines:
\textbf{SD Inpaint + ControlNet-Scribble} \cite{zhang2023adding} guides the diffusion inpainting pipeline using text, scribble, and a mask automatically derived from the scribble.
\textbf{BAGEL-7B-MoT} \cite{deng2025bagel} is a recent unified multimodal model capable of following multimodal image-editing instructions. 

We evaluate both the zero-shot model and a version finetuned on our dataset.

\subsection{Image-Conditioned ControlNet-Scribble}
Inpainting-based methods such as the SD Inpaint baseline require a binary mask that specifies the region to modify. The ControlNet-Scribble model does not condition on the input image and therefore performs image generation rather than editing, often modifying the background. We address these limitations by conditioning ControlNet-Scribble on the latent representation of the original image in addition to text and scribble inputs, enabling scribble-guided editing while preserving the input background.

The original image is encoded with the VAE to produce a 4-channel latent, which is concatenated with the noisy target latent and fed into the diffusion U-Net. We expand the first U-Net convolution from 4 to 8 input channels, initializing the added weights to zero to preserve the pretrained behavior. Text and scribble conditioning follow the standard ControlNet-Scribble implementation.
% \paragraph{Architecture.} Our model extends the U-Net backbone of the diffusion model to incorporate three conditioning inputs: the original image, a text instruction, and a scribble.
% \begin{enumerate}
% \item\textbf{Original image} is encoded with the VAE encoder to produce a 4-channel latent representation. This latent is then concatenated with the 4-channel noisy target latent, forming an 8-channel tensor as U-Net input. We accordingly widen the first convolution layer of the U-Net from 4 to 8 input channels to ingest this concatenated tensor. The newly added kernels are zero-initialized, so the network behaves like the pre-trained model at iteration 0. During training, these weights gradually learn to incorporate the original image information.

% \item\textbf{Text conditioning} is encoded by a pretrained text encoder (e.g., CLIP \cite{radford2021learning}) into a sequence of context embeddings, which are injected into the U-Net through cross-attention layers at each denoising step. This lets the model attend to textual cues when predicting the denoised image latent.

% \item\textbf{Sketch conditioning} is combined with the time embedding and fed into a down-sampling replica of the U-Net encoder. Its convolution blocks produce residual feature maps that are added to the matching U-Net activations. All ControlNet convolutions are zero-initialized, so the sketch has no influence at iteration 0. This ensures the model’s initial output and stability are preserved while learning the new conditioning.
% \end{enumerate}

We train at resolution $512\times512$ with effective batch size 16 using the standard diffusion noise prediction objective. Training proceeds in two stages: we first update the newly introduced layers for 3 epochs with learning rate $1\times10^{-4}$, and then fine-tune the full U-Net for 7 epochs with a smaller learning rate ($1\times10^{-5}$). Optimization uses AdamW with weight decay $1\times10^{-2}$ and gradient clipping. Text and scribble conditions are independently dropped with probability $0.1$ to enable classifier-free guidance (CFG) \citep{ho2022classifier}. At inference, we use 50-step DDIM sampling \citep{song2020denoising} with guidance scale $s=7.5$.

% We minimize the standard $\ell_2$ noise-prediction loss
% $\mathcal{L}_{\text{simple}} = \|\hat\varepsilon - \varepsilon\|_2^2$
% at random diffusion timestep~$t$. Training is carried out in two stages. Stage 1 updates only the $8{\rightarrow}64$ input convolution and all ControlNet weights, while the original SD U--Net is frozen. Stage 2 unlocks the full U-Net but applies a learning rate $10\times$ smaller than that used for the new layers. Text and sketch are each dropped with probability~$0.1$. This teaches the network unconditional paths and enables classifier-free guidance (CFG) \cite{ho2022classifier} at inference.

% \paragraph{Inference}
% At inference, we run DDIM \cite{song2020denoising} with $50$ steps.  For every timestep, we compute both conditional and unconditional predictions and apply CFG:
% $$
%   \hat\varepsilon =
%   \varepsilon_{\text{unc}} + s
%   (\varepsilon_{\text{cond}}-\varepsilon_{\text{unc}}),
% $$
% where a single scale $s$ controls the strength of both text and sketch.

\subsection{BAGEL} We also consider zeroshot and finetuned BAGEL-7B-MoT as alternatives to the finetuned ControlNet-Scribble model. To adapt BAGEL for scribble-guided image-editing, we utilize LoRA finetuning on the language-model backbone. We set LoRA rank to 32 and alpha to 64. We set learning rate to $1\times10^{-4}$ and use a cosine decay, training for 5 epochs with effective batch size 64. To support CFG, we drop the text or sketch inputs with probability 0.1, matching the ControlNet training implementation. We use a spatially weighted MSE loss where we assign the tokens within the target edit area 5 times the weight of the background tokens to encourage the model not to edit the background of images:
\begin{align*}
    \mathcal{L} &= \frac{\sum_{i=1}^{N} (\mathrm{MSE}_i \cdot w_i)}{\sum_{i=1}^{N} w_i}, \\
    w_i &=
    \begin{cases}
        5.0, & \text{if token } i \text{ within target edit mask}, \\
        1.0, & \text{otherwise (background tokens)}.
    \end{cases}
\end{align*}
During inference, we also use 50 steps to match the ControlNet usage.

\begin{figure*}[t]
    \centering
    \includegraphics[width=\textwidth]{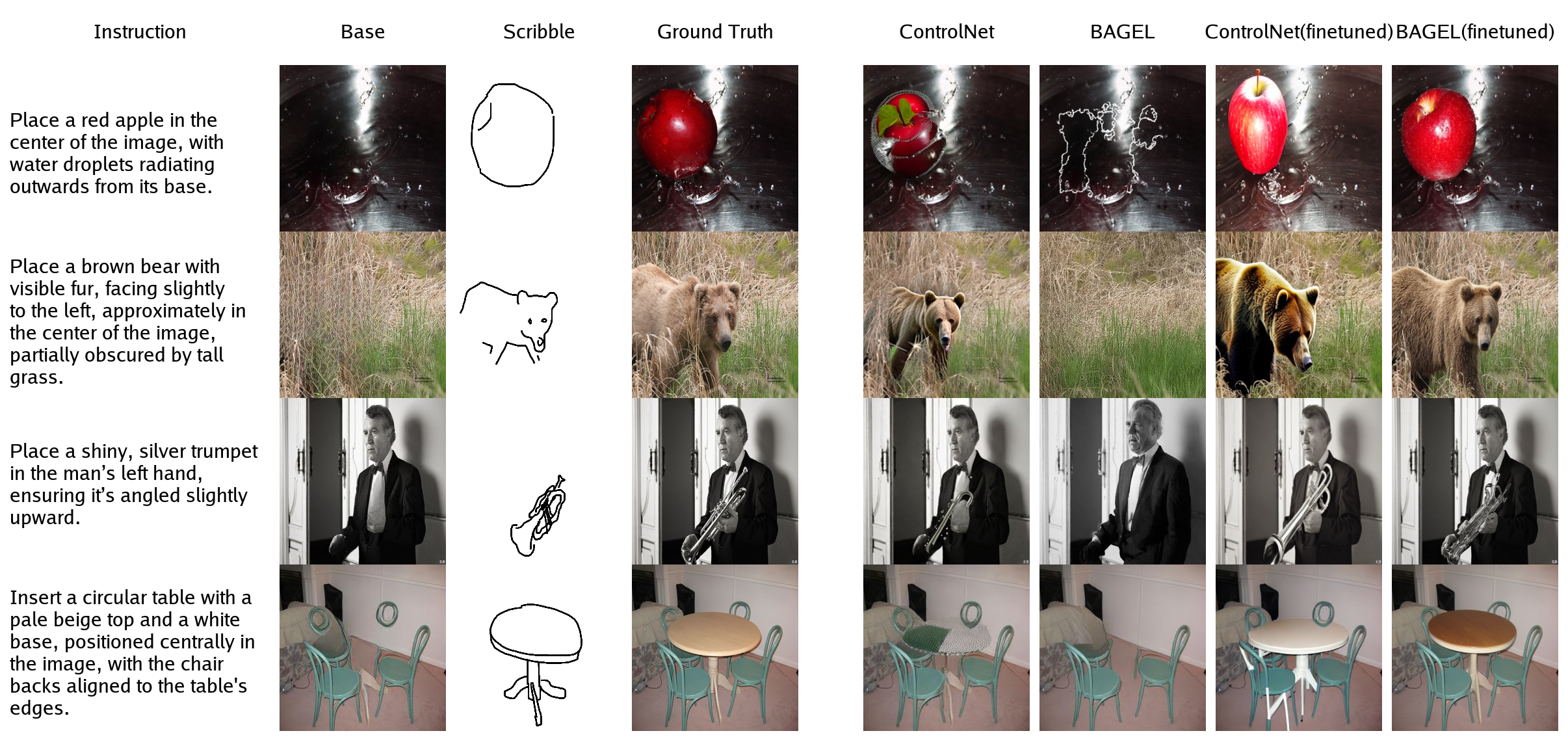}
    \caption{Comparison of editing results across methods. Each row shows one dataset example (Edit Instruction, Original, Scribble, Target) followed by results from four models.}
    \label{fig:qualitative-comparison}
\end{figure*}

\section{Results and Discussion}

We evaluate model outputs using four metrics: Histogram Intersection~\cite{pass1999comparing}, RMSE, CLIP Similarity~\cite{radford2021learning}, and FID~\cite{heusel2017gans}. Histogram Intersection and RMSE are computed both over the entire image and within the bounding box (BBOX) of the edited object. Full results can be found in Table~\ref{tab:editing-results}. Figure~\ref{fig:qualitative-comparison} shows a qualitative comparison of results.

Our results demonstrate that finetuning models on our synthetic dataset results in improvements on across four metrics. Image-conditioned ControlNet-Scribble achieved the lowest FID, indicating that the image conditioning effectively improved realism and preserved background structure. However, weaker performance on other metrics suggests the difficulty of mask-free editing and a trade-off between edit strength and image fidelity.

Before finetuning, BAGEL often generates scribble-like features or even fails to edit the image entirely (see Figure~\ref{fig:qualitative-comparison}), demonstrating that scribble-guided editing may be out of distribution. However, finetuning on our dataset improves results on histogram intersection and RMSE at both the image and bounding-box level, demonstrating increasing similarity at the color and pixel-level. We also see major improvements in CLIP and FID score, which indicates that generations are of higher quality and closer in content to the ground truth images. We find that due to the CFG during training, finetuned BAGEL has minimal performance drops when given only the scribble or text at inference time; only the FID score is significantly affected, rising from 33.36 to 56.96 when using text only. 

As finetuned BAGEL is the best performing model, we perform a small-scale error analysis on fifty images from the test set. We find that 16\% of examples have unclear or incorrect scribble instructions and 34\% have unclear or incorrect text instructions. Of the examples that have correct scribble and text guidance, finetuned BAGEL sufficiently adheres to the scribble in 82\% of examples, but only sufficiently adheres to the text instruction in 57\%. Failures to adhere to scribble guidance typically aligns with a difference in orientation, location, or size from the given scribble. The majority of failures to adhere to text instructions are due to generating an object of incorrect color or the wrong object altogether. It is possible these failure cases can be mitigated with guidance-scale tuning at inference time.

\section{Conclusion}
We introduce \textbf{ScribbleEdit}, a synthetic dataset for multimodal image editing that combines freehand scribbles with natural language instructions. The dataset provides paired supervision for studying scribble-guided editing, where models must interpret abstract spatial cues while preserving the original image context. 
Through experiments with diffusion-based models and multimodal large language models, we show that pretrained models struggle to perform edits from scribble inputs. Training on ScribbleEdit effectively improves their ability to generate spatially aligned and semantically consistent edits.
We hope our dataset and preliminary model encourage future work in precise and human-centered editing tools.

\section{Limitations and Future Directions}
Our dataset currently relies on existing scribbles from the Sketchy Database rather than synthesized scribbles. While this ensures realistic human drawings, it limits the scalability of the synthetic data pipeline. Generating semantically meaningful scribbles remains challenging. Prior work often generates synthetic scribbles using edge detectors (e.g., HED) with heuristic augmentations for training models such as ControlNet-Scribble, which may not fully capture the variability of human drawing behavior. CLIPasso \cite{vinker2022clipasso} suggests promising directions for generating more semantically aware scribbles that could expand our dataset.

In addition, our current dataset focuses on the object addition task, where a removed object is restored using a scribble and text instruction. Future work could explore different types of edits, such as more fine-grained, partial editing. Hierarchical segmentation provided by UnSAMv2 \cite{yu2025unsamv2} could enable constructing datasets with edits at multiple levels of spatial granularity.

Finally, the quality of the dataset depends on automated object removal and instruction generation, which may occasionally introduce noisy examples. Improving dataset filtering and incorporating human evaluation would further strengthen the reliability of the dataset.

{
    \small
    \bibliographystyle{ieeenat_fullname}
    \bibliography{main}
}


\end{document}